\documentclass[a4paper,fleqn]{cas-dc}

\usepackage{booktabs}
\usepackage{helvet}  % DO NOT CHANGE THIS
\usepackage{courier}  % DO NOT CHANGE THIS
\usepackage{amsthm}
\usepackage{xcolor}
\usepackage{graphicx} % DO NOT CHANGE THIS
\usepackage{caption} % DO NOT CHANGE THIS AND DO NOT ADD ANY OPTIONS TO IT
\usepackage{algorithm}
\usepackage{algorithmic}
\usepackage{newfloat}
\usepackage{listings}
\usepackage{amssymb}
\usepackage{amsmath}
\newcommand{\vect}[1]{\boldsymbol{\mathbf{#1}}}

\usepackage{thmtools, thm-restate}
\usepackage[authoryear,longnamesfirst]{natbib}

\def\tsc#1{\csdef{#1}{\textsc{\lowercase{#1}}\xspace}}
\tsc{WGM}
\tsc{QE}
\newtheorem{assumption}{Assumption}
\newcommand{\iid}{i.i.d.~}

\begin{document}
\let\WriteBookmarks\relax
\def\floatpagepagefraction{1}
\def\textpagefraction{.001}

\shorttitle{LSA for Online ZSL with CLIP}    

\shortauthors{}  

\title [mode = title]{Label Shift Aware Adaptation for Online Zero-shot Learning with Contrastive Language-Image Pre-Training (CLIP)}  

\tnotemark[1] 



\author[1]{Pengxiao Han}%[<options>]

% \cormark[1]

\fnmark[1]

% \ead{}


\credit{}

\affiliation[1]{organization={Australian National University},
            city={Canberra},
%            citysep={}, % Uncomment if no comma needed between city and postcode
            state={ACT},
            country={Australia}}

\author[2]{Changkun Ye}%[]

% Footnote of the second author
\fnmark[1]

% Email id of the second author
% \ead{}


\credit{}

\affiliation[2]{organization={China North Vehicle Research Institute},
            city={Beijing},
%            citysep={}, % Uncomment if no comma needed between city and postcode
            country={China}}

% \cortext[1]{Corresponding author}


\author[3]{Yanshuo Wang}%[]

% Footnote of the second author

% Email id of the second author
% \ead{}


\credit{}

\affiliation[3]{organization={The Hong Kong Polytechnic University},
            city={Hong Kong SAR},
%            citysep={}, % Uncomment if no comma needed between city and postcode
%            state={QLD},
            country={China}}

% \cortext[1]{Corresponding author}


\author[1,4]{Jinguang Tong}%[]

% Footnote of the second author

% Email id of the second author
% \ead{}


\credit{}

%\affiliation[1]{organization={},
%            citysep={}, % Uncomment if no comma needed between city and postcode
%            state={QLD},

% \cortext[1]{Corresponding author}


\author[3]{Miaohua Zhang}%[]

% Footnote of the second author

% Email id of the second author
% \ead{}


\credit{}

\affiliation[3]{organization={Griffith University},
            city={Brisbane},
%            citysep={}, % Uncomment if no comma needed between city and postcode
            state={QLD},
            country={Australia}}

% \cortext[1]{Corresponding author}


\author[4,1]{Xuesong Li}%[]

% Footnote of the second author

\cormark[1]

% Email id of the second author
\ead{xuesong.li@csiro.au}


\credit{}

\affiliation[4]{organization={CSIRO},
            city={Canberra},
%            citysep={}, % Uncomment if no comma needed between city and postcode
            state={ACT},
            country={Australia}}


\author[5]{Jie Hong}[orcid=0009-0000-3855-535X]

% Footnote of the second author

\cormark[1]

% Email id of the second author
\ead{jiehong@hku.hk}


\credit{}

\affiliation[5]{organization={The University of Hong Kong},
            city={Hong Kong SAR},
%            citysep={}, % Uncomment if no comma needed between city and postcode
            country={China}}

\author[4]{Lars Petersson}%[]

% Footnote of the second author

% Email id of the second author
% \ead{}


\credit{}

% \affiliation[1]{organization={},
% %           citysep={}, % Uncomment if no comma needed between city and postcode

\fntext[1]{Equal contributions}

\cortext[1]{Corresponding author}


\begin{abstract}
Vision-language models like Contrastive Language-Image Pre-Training (CLIP) have been extensively studied in data-scarce scenarios. A particularly challenging and realistic task in this area is \textit{online zero-shot learning with CLIP}, where unknown test samples are predicted sequentially in random order by CLIP while keeping the feature extraction and model parameters fixed during the sequential inference phase. Most existing approaches in this setting address the problem by adapting representations online using incoming test samples, while neglecting the distribution of the data on which CLIP was initially trained. This mismatch can lead to degraded performance when the label distribution in the test data differs from that of the training domain. To address this gap, we propose Label Shift Aware (LSA), which formulates the online zero-shot classification task as a domain adaptation problem. Specifically, LSA adapts the predictions computed by CLIP, which was trained on an unknown source distribution, to a target distribution using only unlabeled test data, and applies label shift correction to mitigate the mismatch between the source and target domains. The extensive experiments across multiple datasets demonstrate that the proposed LSA consistently outperforms state-of-the-art online zero-shot learning methods based on CLIP.
\end{abstract}


%\nocite{*}

% Keywords

\maketitle

\section{Introduction}
\label{sec:intro}
The rise of foundation models, including large language models (LLMs)~\citep{chatgpt}, large vision models (LVMs)~\citep{sam_kirillov2023segany, dino_caron2021emergingpropertiesselfsupervisedvision}, and Vision-language models (VLMs)~\citep{sora, sd_rombach2022high}, has significantly advanced performance across a wide range of computer vision and machine learning tasks~\citep{Dreamfusion_poole2022dreamfusion, textinv_gal2022image, Instructpix2pix_brooks2023instructpix2pix,hu2025inference}. Among these, CLIP (Contrastive Language–Image Pretraining) has attracted particular attention for its strong generalization across unseen visual concepts. This open-vocabulary recognition capability enables CLIP to perform well without further fine-tuning in two challenging settings: Zero-Shot Learning (ZSL) and Test-Time Adaptation (TTA), which are increasingly studied as complementary approaches for addressing different types of generalization challenges, particularly those arising from distribution or domain shift.

\begin{figure*}[t]
    \centering
    \includegraphics[width=1.0\linewidth]{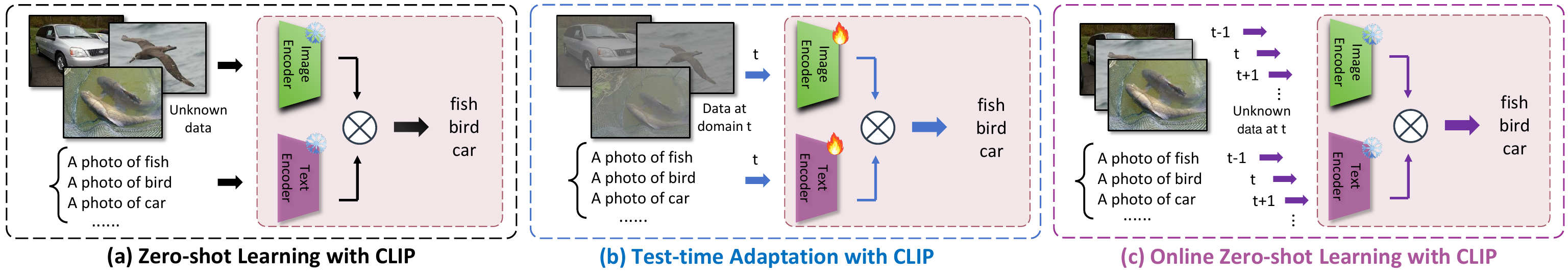}
    \caption{Comparison of three CLIP-based inference settings. (a) ZSL with CLIP: The model makes predictions on unseen data using pre-trained text and image encoders, without any adaptation. (b) TTA with CLIP: The model adapts using access to multiple test samples from the target domain. (c) Online ZSL with CLIP: Images arrive sequentially in a streaming manner, and each image is predicted immediately without storage.} 
    \vspace{-12pt}
    \label{fig:intro}
\end{figure*}

In ZSL with CLIP, as shown in Figure~\ref{fig:intro} (a), CLIP classifies images from previously unseen categories without accessing any labeled examples of these unseen classes. This is typically achieved by converting class names into textual prompts (\textit{e.g.}, "a photo of a dog") and computing the similarity between the image and text embeddings. Owing to its large-scale pretraining on massive image-text pairs, CLIP generalizes efficiently to novel categories. While CLIP demonstrates impressive zero-shot performance, its predictions can be sensitive to the phrasing of textual prompts and the quality of image representations. Therefore, many approaches aim to improve prompt quality, either by tuning textual prompts during inference~\citep{tpt_shu2022testtimeprompttuningzeroshot}, enriching descriptions with external knowledge~\citep{adaptCLIP_saha2024improvedzeroshotclassificationadapting}, or aligning the token-level distribution between modalities~\citep{PromptAlign_hassan2024alignpromptstesttimeprompting}. Complementary to this, other methods focus on enhancing visual representations, for example, by clustering image features to construct better visual proxies~\citep{InMap_qian2023intramodalproxylearningzeroshot}, or by introducing lightweight parameter adaptation modules to guide prediction confidence~\citep{TTL_imam2024testtimelowrankadaptation}. Unlike ZSL, which improves CLIP's robustness by enhancing prompt quality or visual representation under aligned distribution, TTA with CLIP aims to extend this robustness when the test distribution differs from the distribution seen during pretraining (see Figure~\ref{fig:intro} (b)). TTA leverages unlabeled test data during inference to refine representations or fine-tune models, enabling adaptation without access to training labels. Representative works include test-time prompt optimization~\citep{Dart_liu2024dart, SwapPrompt_NEURIPS2023_cdd06402}, augmentation-driven feature refinement~\citep{MTA_zanella2024test}, lightweight adaptation modules~\citep{TTL_imam2025test}, and filtering or denoising test samples~\citep{AdaND_cao2025noisy}.

Building on the progress of ZSL and TTA, a more challenging and realistic setting has recently emerged: online ZSL with CLIP~\cite{onzeta_qian2024online}, which is illustrated in Figure~\ref{fig:intro} (c). This setting not only addresses the classification of previously unseen classes but also operates under a streaming scenario where data arrives sequentially. Specifically, online learning in this context refers to a sequential, single-pass setup: each instance in the data stream is predicted and then used to update the model, without storing past data. This constraint reflects real-world limitations such as data privacy, memory constraints, and computational efficiency, making it especially relevant for deployment on edge devices, robotics, or mobile platforms. The key characteristics of this online setting are that each instance is seen only once in a single pass, no data can be stored for future use, the model must make real-time predictions, and it must do so without consuming excessive computational resources. Prior works, such as OnZeta~\citep{onzeta_qian2024online} and TPT~\citep{tpt_shu2022testtimeprompttuningzeroshot}, have primarily focused on improving multi-modal alignment at test time and refining prompt-based semantic representations. 

Despite the effectiveness of prior approaches, they often overlook the label distribution shift between the training data used to pretrain CLIP and the target data encountered in real-world online streams. This mismatch can lead to biased predictions and degraded performance in online zero-shot settings. To address this challenge, we propose Label Shift Aware, a novel approach that integrates dynamic label distribution estimation into the online zero-shot pipeline. Specifically, LSA continuously estimates the evolving label distribution from the incoming test stream using model predictions and adjusts the classifier's decision scores via label prior reweighting. This approach allows LSA to correct bias induced by the mismatched label between the source and target domains. By adapting to the test-time label distribution without storing past data or retraining CLIP, LSA offers a lightweight, memory-efficient solution that is well-suited for streaming scenarios.
We validate the effectiveness of our method across 14 datasets, demonstrating consistent performance improvement. 
The contributions of this work can be summarized as follows:
\begin{itemize}
    \item We formulate online ZSL with CLIP as a label shift problem, highlighting that the evolving label distribution in the test stream is a major cause of performance degradation in online settings. This formulation offers a new perspective on the task and underscores the need for dynamic adaptation.
    \item We propose a non-parametric, memory-efficient estimator that dynamically tracks the test-time label distribution using model predictions, without storing past data or requiring ground-truth labels.
    \item We introduce a posterior adjustment mechanism that reweights CLIP’s predicted class probabilities based on the estimated label priors, enabling adaptation without modifying the backbone model.
    \item The proposed method is lightweight, modular, and model-agnostic, allowing easy integration with any CLIP-based zero-shot classifier for real-time deployment in streaming environments.
\end{itemize}

\section{Related Works}
\label{sec:related_work}
\subsection{Zero-shot Learning with CLIP}
Recently, several works have leveraged vision-language models such as CLIP~\cite{clip_radford2021learning} to tackle ZSL~\citep{adaptCLIP_saha2024improvedzeroshotclassificationadapting, InMap_qian2023intramodalproxylearningzeroshot, PromptAlign_hassan2024alignpromptstesttimeprompting, tpt_shu2022testtimeprompttuningzeroshot, TTL_imam2024testtimelowrankadaptation}.
Test-Time Prompt Tuning (TPT)~\citep{tpt_shu2022testtimeprompttuningzeroshot} improves zero-shot generalization of CLIP by tuning the text prompts during inference, without modifying the model backbone or using labeled training data. 
PromptAlign~\citep{PromptAlign_hassan2024alignpromptstesttimeprompting} addresses the task by introducing a loss function that combines entropy minimization with a token-level distribution alignment loss, which encourages images and text prompt tokens to follow a similar distribution. Both TPT and PromptAlign augment the test sample to ensure stability and semantic consistency. AdaptCLIP~\citep{adaptCLIP_saha2024improvedzeroshotclassificationadapting} enriches class representations by retrieving natural language descriptions from external sources (\textit{e.g.}, Wikipedia), which are then used as input prompts. Based on the class names, each test image is then randomly paired with a textual description generated by an LLM. 
InMap~\citep{InMap_qian2023intramodalproxylearningzeroshot} improves zero-shot classification by improving alignment within the visual modality of CLIP. It learns visual proxies for each class by clustering image features from unlabelled test data, which provides a better representation of class semantics in the visual feature space.
TTL~\citep{TTL_imam2024testtimelowrankadaptation} injects trainable low-rank matrices (LoRA-style) into the text encoder and adapts them only at test time, guided by a confidence maximization objective. The method encourages the model to make more confident predictions on unlabeled test data without modifying the pre-trained model weights.

\subsection{Test-time Adaptation with CLIP}
CLIP has shown remarkable improvements in ZSL. At the same time, recent works have explored its performance under test-time adaptation (TTA) settings~\citep{Dart_liu2024dart, MTA_zanella2024test, wang2025dynamic, wang2024maintain, wang2024backpropagation, han2024latent}. 
TTA predicts test data on multiple new target domains. Prior to testing, the model has been pre-trained on source-domain data. During testing, the model predicts data in the target domain and adjusts its parameters using data from the next domain.
DART~\citep{Dart_liu2024dart} introduces a dual-modal prompting mechanism where learnable prompts are optimized online using unlabeled test data. This reduces predictive uncertainty and enhances alignment between image and text modalities. Maxime \textit{et al.}~\citep{MTA_zanella2024test} propose MeanShift Test-Time Augmentation (MTA) that applies mode seeking directly in the visual embedding space. By leveraging multiple augmented views of the same image, MTA achieves strong performance without requiring access to gradients, training, or hyperparameter tuning. RLCF~\citep{RLCF_zhao2023test} criticizes the overconfidence issue introduced by entropy-based methods like TPT, treating CLIP as a reward model, and applying reinforcement learning at test time to guide the VLM toward a better predictive distribution. Traditional TTA methods assume that test samples belong to in-distribution classes. However, in open-world settings, test data often includes noisy or unknown samples. To address this, Cao et al.~\cite{AdaND_cao2025noisy} design a lightweight linear detector on top of frozen CLIP image features to distinguish noisy samples, and inject pure Gaussian noise to improve robustness. SwapPrompt~\citep{SwapPrompt_NEURIPS2023_cdd06402} employs a dual-prompt strategy, where an online prompt is updated at each step, and an exponential moving average (EMA) prompt retains historical knowledge. The method uses Prompt Swapped Prediction to introduce a cross-prompt loss, thereby effectively enhancing zero-shot generalization across unseen domains.

\subsection{Online Zero-shot Learning with CLIP}
Online ZSL with CLIP provides a practical setting for real-world scenarios, such as mobile devices and robotics, where models are deployed in resource-constrained environments. In this setting, test samples arrive sequentially as a data stream, cannot be stored, and must be classified immediately without any offline refinement. This makes the classification task more challenging than conventional ZSL with CLIP and TTA with CLIP. To address this, OnZeta~\citep{onzeta_qian2024online} proposes a tailored approach: it first fixes both the vision and text proxies and estimates the label distribution of the target stream. Then, the method aligns the CLIP text proxy to the visual proxy, thereby reducing the modality gap. Experimental results show that OnZeta is highly effective for online ZSL with CLIP, achieving strong performance under this challenging setting. In this work, we focus on addressing online ZSL with CLIP by modeling the evolving label distribution in the test stream and correcting the mismatch between the training and deployment domains through dynamic label-shift adaptation.

\section{Methodology}
In this section, we propose the Label Shift Aware framework to address the challenge of online ZSL with CLIP under distributional shift. LSA dynamically adapts the classifier by modeling the evolving test-time label distribution, without requiring labeled data or retraining the backbone. The overall framework consists of three key components: LSA Weight Generation, Active Prior Adaptation, and LSA Classifier Correction. These modules work together to estimate the test-time label prior and adjust predictions accordingly, enabling robust and efficient online zero-shot classification.

\begin{figure*}
    \centering
    \includegraphics[width=1.0\linewidth]{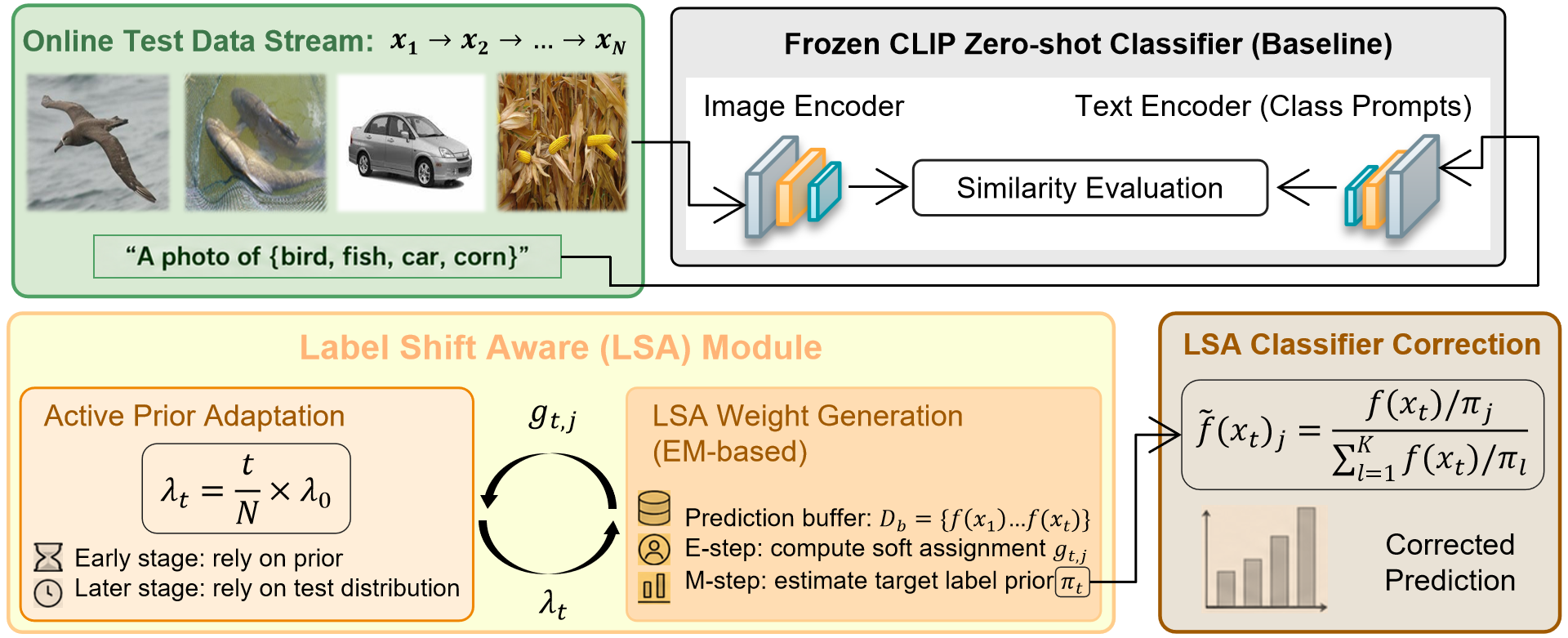}
    \caption{The framework of the proposed LSA. Test samples arrive sequentially in a streaming manner. A frozen CLIP zero-shot classifier produces prediction scores without any parameter updates. LSA estimates the evolving target label distribution using an EM-based label shift estimator with an adaptive prior, and corrects the classifier output via posterior reweighting. The entire framework is training-free, memory-efficient, and suitable for online deployment.} 
    \vspace{-8pt}
    \label{fig:method}
\end{figure*}

\subsection{Preliminaries}
\noindent \textbf{Problem Definition.} In this work, we consider the problem of online zero-shot classification with CLIP. Given a CLIP-based image classifier $f(x)$, we aim to adapt the classifier $f(x)$ to the test data distribution based on unlabeled test samples coming in a stream.

\noindent \textbf{Notations.} We denote $\mathcal{X}\in\mathbb{R}^d$ as the data space, $\mathcal{Y}\in\{1,2,...,K\}$ as the label space of $K$ classes, $f:\mathcal{X}\rightarrow\Delta^{K-1}$ as the CLIP-based classifier that output SoftMax predictions and $\mathcal{D}^{target}=\{x_t\}^N_{t=1}$ as the unlabeled test dataset where each sample $x_t$ appearing at order or time $t$.

\subsection{Motivation Overview}
In this work, as shown in Figure~\ref{fig:method}, we view the online zero-shot classification with CLIP problem as a special type of domain adaptation task. In this sense, the task can be interpreted as adapting the classifier $f$ trained on an unknown source domain or train distribution $p_{source}(x,y=\cdot)$ to the target domain or test distribution $p_{target}(x,y=\cdot)$, given on-the-fly target unlabeled data from $\mathcal{D}^{target}=\{x_t\}^N_{t=1}$ with $x_t\sim_{\iid}p_{target}(x)$.
In domain adaptation tasks, covariate shift and label shift are two main types of distribution shift~\cite{RLLS,alexandari2020maximum,osls}.
When deploying a model trained on the source domain to a target domain, covariate shift occurs when the data distribution $p(x)$ shifts while $p(y|x)$ remains invariant. Label shift happens when the label distribution $p(y)$ is shifted while $p(x|y)$ is invariant. Both the covariate and label shifts between the two domains can lead to sub-optimal model performance in the target domain~\cite{garg2020unified}. 

Tackling covariate and label shift together can be challenging. In this work, we focus on tackling label shift in online ZSL using CLIP. Since large models (\textit{e.g.}, CLIP) are usually trained with samples from a variety of data sources, the source data distribution $p_{source}(x)$ could effectively cover the target data distribution $p_{target}(x)$, thereby mitigating the problem of covariate shift. In this sense, label shift could be the main factor that impacts the performance of the CLIP-based classifier in our active-ZSL task.

\subsection{Label Shift Aware Weight Generation}
We use $p_{\text{source}}(x,y=\cdot)$ and $p_{\text{target}}(x,y=\cdot)$
to denote the train and test data distributions, respectively.
Without loss of generality, the train and test labels follow
categorical distributions:
\begin{align}
Y_{\text{source}} \sim \text{Cat}(K,\vect{c}), \quad
Y_{\text{target}} \sim \text{Cat}(K,\vect{\pi}),
\end{align}
and
\begin{align}
\qquad \qquad p_{\text{source}}(y=\cdot) = \vect{c} \in \Delta^{K-1}, \\
\qquad \qquad p_{\text{target}}(y=\cdot) = \vect{\pi} \in \Delta^{K-1},
\end{align}
where $\Delta^{K-1}$ denotes the $K$-dimensional probability simplex.
In the online zero-shot classification tasks, we are given with a source domain classifier $f:\mathcal{X}\rightarrow\Delta^{K-1}$ and a target domain unlabeled dataset $\mathcal{D}^{target}=\{x_t\}^N_{t=1}$ with each sample ${x_t\sim_{i.i.d.}p_{target}(x)}$ coming in stream.

\begin{algorithm}[t]
\caption{\textbf{LSA Weight Generation}}
	\begin{algorithmic}
	    \label{alg:LAS}
	    \STATE \textbf{Input: } 
            \begin{itemize}
                \item Test prediction buffer $\mathcal{D}^b=\{f(x_t)\}^{N}_{t=1}$ at time $t=1,2,...,N$;
                \item CLIP-based classifier $f(x)$;
                \item Prior information $\vect{\alpha}\in\mathbb{R}^K_{>1}$.
            \end{itemize}
	    \STATE \textbf{Initialize: } $\vect{\pi}^{(0)}\in \Delta^{K-1}_{>0}$.
		\FOR{$m=0$ to $M$}
        \STATE \textbf{E-step} Evaluate $g^{(m)}_{t,j}$:
        \STATE\begin{equation}
                \label{alg-eq:E}
                g^{(m)}_{t,j} = \frac{\pi^{(m)}_j\cdot f(x_t)_j}{\sum^K_{l=1}\pi^{(m)}_l\cdot f(x_t)_l},
            \end{equation}
        \STATE  \textbf{M-step} Obtain $\vect{\pi}^{(m+1)}$ with:
        \STATE \begin{equation}
                \label{alg-eq:M}
                        \pi_j^{(m+1)} = \lambda_t\cdot \frac{\sum^{t}_{i=1} g^{(m)}_{i,j}}{N} + (1-\lambda_t)\cdot \frac{\alpha_j-1}{\sum^K_{l=1}(\alpha_l - 1)},
                        % \pi_j^{(m+1)} = \lambda\cdot \frac{\sum^{N}_{i=1} g^{(m)}_{i,j}}{N} + (1-\lambda)\cdot \frac{\alpha_j-1}{\sum^K_{l=1}(\alpha_l - 1)}
                \end{equation}
		\ENDFOR
		\STATE \textbf{Output: } $\vect{\pi} = \vect{\pi}^{(M+1)}$
	\end{algorithmic}
\end{algorithm}

We propose using a label-shift estimation method to address the online zero-shot classification task. The label shift assumption can be written as:
\begin{assumption}\label{assume:a1}
    (\textbf{Label Shift Assumption}~\cite{BBSE})
\[
    p_{source}(x|y=j) = p_{target}(x|y=j) \quad \text{for all} \quad j\in\mathcal{Y}.
\]
\end{assumption}

If we can estimate $\vect{\pi}$ and $\vect{c}$, Given the streaming test dataset $\mathcal{D}^{target}$ and the classifier $f$, we have:
\begin{restatable}{lemma}{nll}
Under Assumption~\ref{assume:a1}, the negative log likelihood of $\vect{\pi}$ and $\vect{c}$ can be written as follows:
\begin{equation}\label{eq:nll}
        -\log L(\vect{\pi},\vect{c};\mathcal{D}^{target}) =  -\sum^N_{t=1}\log \sum^K_{j=1}\frac{\pi_j}{c_j} f(x_t)_j - C,
\end{equation}
where $C$ does not depend on $\vect{\pi}$ or $\vect{c}$.
\end{restatable}
Moreover, with Bayesian inference methods, we can employ a Dirichlet prior $p(\vect{\pi}\mid\vect{\alpha})\sim \text{Dir}(K,\vect{\alpha})$ over the target label distribution $\vect{\pi}$, we can construct the posterior of $\vect{\pi}$ given prior and dataset $\mathcal{D}^{target}$ based on the negative log likelihood~\cite{mapls}. The posterior can thus be written as:
\begin{equation}\label{eq:posterior}
    p(\vect{\pi}\mid\mathcal{D}^{target},\vect{\alpha}) = \frac{1}{C}\cdot\prod^K_{l=1}\pi_l^{\alpha_{l}-1} \cdot \prod^N_{t=1} \sum^K_{j=1}\frac{\pi_j}{c_j} f(x_t)_j,
\end{equation}
where $\alpha_l>1$ for $l=1,2,..,K$ is the element of the $K$-dimensional parameter $\vect{\alpha}$ of the Dirichlet prior and $C$ is a parameter irrelevant to $\vect{\pi}$ or $\vect{c}$.

Under the label shift problem setup, the MAPLS algorithm proposed in \cite{mapls} estimates the target label distribution $p_{target}(y=\cdot)=\vect{\pi}$ through optimizing the negative log likelihood Equation~\eqref{eq:nll} with respect to the parameter $\vect{\pi}$. However, this model cannot be directly applied in our task due to the unknown parameter $\vect{c}$ and the on-the-fly test samples under the active ZSL setup.

In this work, we utilize the MAPLS algorithm to propose the EM algorithm in Algorithm~\ref{alg:LAS} to estimate the target label distribution $\vect{\pi}$. As shown in Algorithm~\ref{alg:LAS}, we postulate a uniform source label distribution $\vect{c}:=\vect{1}/K$, expecting that the large model is roughly the same for different classes.

In Algorithm~\ref{alg:LAS}, the E-step and the M-step are evaluated alternately to obtain the final estimate. In the M-step, the $\vect{\pi}^{(m+1)}$ is calculated by a linear combination of the data term with $g_{ij}$ and the Dirichlet prior term with $\vect{\alpha}$. The hyperparameter $\lambda_0$ balances the trade-off between the contribution of the two information to the final estimation, which is defined as:
\begin{equation}\label{eq:lambda0}
\qquad \qquad \lambda_0 = \frac{N}{N +  \sum^K_{l=1}(\alpha_l-1)},
\end{equation}
where $\lambda_0, \in (0,1]$, is decided by the prior information $\vect{\alpha}$.
A higher value of $\lambda_0$ gives greater weight to the test-time label distribution during LSA weight generation, making the model more adaptive to the test stream. Conversely, a lower $\lambda_0$ retains more influence from the pre-defined training prior. The ablation study of $\lambda_0$ is conducted in the experimental part of the supplementary material.

\begin{algorithm}[t]
\caption{\textbf{Overall LSA Model}}
	\begin{algorithmic}
	    \label{alg:LAS-all}
	    \STATE \textbf{Input:} 
            \begin{itemize}
                \item Test data stream $x_t$ at time $t=1,2,...,N$;
                \item CLIP-based classifier $f(x)$;
                \item Prior hyper-parameter $\lambda_0$.
            \end{itemize}
        \STATE \textbf{Initialize:} Prediction buffer $\mathcal{D}^b=\{\}$.
		\FOR{$t=1$ to $N$}
        \STATE Append prediction $f(x_t)$ in buffer $\mathcal{D}^b=\{f(x_t)\}^N_{t=1}$;
        \STATE Update $\lambda_t$ with Equation~\eqref{eq:lambda-schedule};
        \STATE Obtain $\vect{\pi}$ through Algorithm~\ref{alg:LAS};
        \STATE Correct the updated prediction through Equation~\eqref{eq:lsc};
		\ENDFOR
		\STATE \textbf{Output:} Overall classification accuracy comparing $\mathcal{D}^b$ and the ground truth.
	\end{algorithmic}  
\end{algorithm}  

\subsection{Active Prior Adaptation}
In the active learning setting, the target-domain samples arrive in a stream. Therefore, our estimation of the weight could suffer from high estimation error. To mitigate this problem, we propose an adaptive scheme to adjust the prior weight $\lambda$ in the LSA weight estimation algorithm (Algorithm~\ref{alg:LAS}), with the equation as follows:
\begin{equation}\label{eq:lambda-schedule}
\qquad \qquad \qquad \quad \lambda_t = \frac{t}{N}\cdot \lambda_0, 
\end{equation}
where $\lambda_0$ is the initial value of $\lambda$ in Equation~\ref{eq:lambda0} and $t$ is current time. Early in the stream (small $t$), we rely more on prior knowledge since we have limited samples. As $t$ approaches $N$, we trust the empirical distribution more, hence $\lambda$ increases linearly. In this way, the contribution of data to the final estimate of $\vect{\pi}$ will gradually increase, thus enabling prior information to provide more regularization at the early stage.

\subsection{Label Shift Aware Classifier Correction}
Getting the estimated weight $\vect{\pi}$ from Algorithm~\ref{alg:LAS}, we propose to adjust the prediction of the model at time $t$ with the following equation:
\begin{equation}\label{eq:lsc}
\qquad \quad \Tilde{f}(x_t)_j = \frac{f(x_t)/\pi_j}{\sum^K_{l=1}f(x_t)/\pi_l},
\end{equation}
where the estimated $\vect{\pi}$ corrects the prediction output~\cite{xu2021towards}. Intuitively, the estimated $\vect{\pi}$ attempts to compensate for the discrepancies between the distributions of training data of CLIP and the online test data. 

\subsection{Overall Framework}
After correcting the CLIP-based classifier $f$, the overall framework of the LSA model for the online ZSL with CLIP can be summarized in Algorithm~\ref{alg:LAS-all}. The core of the algorithm is to generate the weight vector $\vect{\pi}$ using label shift to adjust the classifier's predictions over time. During test time, $\lambda$, which balances the influence of the test data and the training prior, is made adaptive. The LSA model does not update its weights. In addition to its training-free property, it can be compatible with any baselines. In the experimental section, we apply LSA using the CLIP and OnZeta as the baselines.

\section{Experiments}
In this section, we evaluate the effectiveness of the proposed LSA in the online ZSL setting with CLIP. Experiments are conducted on 14 diverse datasets covering general classification, fine-grained recognition, and domain-shifted scenarios, enabling a comprehensive assessment of LSA’s generalization capabilities. We compare our method against state-of-the-art baselines. To better understand the contributions of our framework, we conduct ablation studies in two parts: the first evaluates the individual contributions of each core component; the second examines the sensitivity of LSA to the hyperparameter that controls the influence of label shift correction during inference. All results are reported using top-1 accuracy and follow the evaluation protocol introduced in \cite{onzeta_qian2024online} to ensure fair comparison.
More experimental settings are provided in the supplementary material.

\begin{table*}[t]
\centering
\resizebox{1.0\textwidth}{!}{
\begin{tabular}{lcccccccccccccc}
\toprule
\textbf{Dataset} &\textbf{Aircraft} &\textbf{Caltech} &\textbf{Cars} &\textbf{Cifar10} &\textbf{Cifar100} &\textbf{CUB} &\textbf{DTD} &\textbf{EuroSAT} &\textbf{Flowers} &\textbf{Food} &\textbf{Pets} &\textbf{SUN} &\textbf{UCF101} &\textbf{Avg.} \\
\midrule

&\multicolumn{14}{c}{\textbf{ResNet-50}} \\ \midrule
\textbf{CLIP baseline} &16.92 &79.14 &54.27 &71.58 &41.91 &42.29 &42.39 &31.60 &65.69 &80.59 &84.23 &52.59 &59.81 &55.62 \\
\textbf{TPT}           &17.58 &\textbf{87.02} &58.46 &–     &–     &–     &42.33 &28.83 &63.12 &74.89 &81.34 &61.46 &56.60 &– \\
\textbf{OnZeta}        &17.26 &79.39 &54.55 &71.57 &46.36 &42.46 &41.73 &29.94 &64.90 &80.85 &83.96 &53.27 &60.81 &55.93 \\ \midrule
\textbf{LSA (ours)}    &18.12 &78.82 &56.56 &77.54 &46.53 &46.01 &44.41 &40.30 &66.86 &81.58 &85.70 &52.88 &60.60 &58.15 \\
\textcolor{purple}{Improvements} &\textcolor{purple}{$\uparrow$1.20} &\textcolor{purple}{$\downarrow$0.32} &\textcolor{purple}{$\uparrow$2.29} &\textcolor{purple}{$\uparrow$5.96} &\textcolor{purple}{$\uparrow$4.62} &\textcolor{purple}{$\uparrow$3.72} &\textcolor{purple}{$\uparrow$2.02} &\textcolor{purple}{$\uparrow$8.70} &\textcolor{purple}{$\uparrow$1.17} &\textcolor{purple}{$\uparrow$0.99} &\textcolor{purple}{$\uparrow$1.47} &\textcolor{purple}{$\uparrow$0.29} &\textcolor{purple}{$\uparrow$0.79} &\textcolor{purple}{$\uparrow$2.53}    \\
\textbf{LSA+OnZeta (ours)}          &\textbf{18.88} &79.63 &\textbf{58.89} &\textbf{77.64} &\textbf{48.36} &\textbf{46.63} &\textbf{44.81} &\textbf{41.10} &\textbf{66.22} &\textbf{82.01} &\textbf{84.84} &\textbf{53.62} &\textbf{61.54} &\textbf{58.78} \\
\textcolor{purple}{Improvements}    &\textcolor{purple}{$\uparrow$1.96} &\textcolor{purple}{$\uparrow$0.49} &\textcolor{purple}{$\uparrow$4.62} &\textcolor{purple}{$\uparrow$6.06} &\textcolor{purple}{$\uparrow$6.45} &\textcolor{purple}{$\uparrow$4.34} &\textcolor{purple}{$\uparrow$2.42} &\textcolor{purple}{$\uparrow$9.50} &\textcolor{purple}{$\uparrow$0.53} &\textcolor{purple}{$\uparrow$1.42} &\textcolor{purple}{$\uparrow$0.61} &\textcolor{purple}{$\uparrow$1.03} &\textcolor{purple}{$\uparrow$1.73} &\textcolor{purple}{$\uparrow$3.16} \\
\midrule

&\multicolumn{14}{c}{\textbf{ViT-B/16}} \\ \midrule
\textbf{CLIP baseline} &24.36 &83.91 &64.66 &90.77 &68.27 &52.66 &45.27 &41.38 &71.08 &88.86 &87.84 &62.97 &72.43 &65.96 \\
\textbf{TPT}           &24.78 &\textbf{94.96} &\textbf{68.46} &–     &–      & –   &46.35 &33.60 &69.26 &86.97 &\textbf{89.00} &67.21 &58.91 & – \\
\textbf{OnZeta}        &24.81 &83.86 &64.97 &90.97 &70.78 &52.97 &45.02 &40.32 &70.49 &89.07 &87.97 &63.65 &73.31 &66.21 \\ \midrule
\textbf{LSA (ours)}    &26.28 &84.46 &66.43 &91.54 &70.63 &56.33 &45.85 &45.64 &73.82 &89.51 &88.02 &63.12 &72.97 &67.28 \\
\textcolor{purple}{Improvements}      &\textcolor{purple}{$\uparrow$1.92} &\textcolor{purple}{$\uparrow$0.55} &\textcolor{purple}{$\uparrow$1.77} &\textcolor{purple}{$\uparrow$0.77} &\textcolor{purple}{$\uparrow$2.36} &\textcolor{purple}{$\uparrow$3.67} &\textcolor{purple}{$\uparrow$0.58} &\textcolor{purple}{$\uparrow$4.26} &\textcolor{purple}{$\uparrow$2.74} &\textcolor{purple}{$\uparrow$0.65} &\textcolor{purple}{$\uparrow$0.18} &\textcolor{purple}{$\uparrow$0.15} &\textcolor{purple}{$\uparrow$0.54} &\textcolor{purple}{$\uparrow$1.32}           \\ 
\textbf{LSA+OnZeta (ours)}            &\textbf{26.67} &84.72 &68.26 &\textbf{91.85} &\textbf{71.70} &\textbf{57.11} &\textbf{46.88} &\textbf{46.18} &\textbf{73.31} &\textbf{89.85} &88.20 &\textbf{63.93} &\textbf{74.02} &\textbf{67.90} \\
\textcolor{purple}{Improvements}     &\textcolor{purple}{$\uparrow$2.31} &\textcolor{purple}{$\uparrow$0.81} &\textcolor{purple}{$\uparrow$3.60} &\textcolor{purple}{$\uparrow$1.08}  &\textcolor{purple}{$\uparrow$3.43} &\textcolor{purple}{$\uparrow$4.45} &\textcolor{purple}{$\uparrow$1.61} &\textcolor{purple}{$\uparrow$4.80}  &\textcolor{purple}{$\uparrow$2.23} &\textcolor{purple}{$\uparrow$0.99} &\textcolor{purple}{$\uparrow$0.36} &\textcolor{purple}{$\uparrow$0.96}  &\textcolor{purple}{$\uparrow$1.59} &\textcolor{purple}{$\uparrow$1.94} \\

\bottomrule
\end{tabular}}
\caption{Performance comparison among the CLIP baseline, TPT~\cite{tpt_shu2022testtimeprompttuningzeroshot}, OnZeta~\cite{onzeta_qian2024online}, and our proposed LSA method across different datasets. Classification accuracy (in \%) is reported for two backbones: ResNet-50 and ViT-B/16. The best-performing results are highlighted in bold.}
\vspace{-12pt}
\label{tab:dataset}
\end{table*}

\begin{table}[t]
    \centering
    \resizebox{0.44\textwidth}{!}{
    \begin{tabular}{lccc}
        \toprule
        \textbf{Backbone} & \textbf{CLIP baseline} & \textbf{OnZeta} & \textbf{LSA (ours)} \\
        \midrule
        
        \textbf{ResNet-50} &60.30  &62.29  &\textbf{62.86} \textcolor{purple}{($\uparrow$ 2.56)} \\

        \textbf{ViT-B/32}  &63.80  &65.67  &\textbf{66.18} \textcolor{purple}{($\uparrow$ 2.38)} \\
        
        \textbf{ViT-B/16}  &68.81  &70.81  &\textbf{71.30} \textcolor{purple}{($\uparrow$ 2.49)} \\
        
        \textbf{ViT-L/14}  &75.93  &77.72  &\textbf{78.11} \textcolor{purple}{($\uparrow$ 2.18)} \\
        
        \textbf{ViT-L/14$_{336}$} &77.00 &78.73  &\textbf{79.12} \textcolor{purple}{($\uparrow$ 2.12)} \\
        \bottomrule
        
    \end{tabular}}
    \caption{Performance comparison of CLIP baseline, onZeta~\cite{onzeta_qian2024online}, and LSA using different backbones on ImageNet~\citep{imagenet_russakovsky2015imagenet}.}
    % The classification accuracy (in \%) is reported, with the best-performing results being highlighted in bold.} 
    % LSA indicates improvement over the CLIP baseline.}
    \vspace{-12pt}
    \label{tab:imagenet}
\end{table}

\subsection{Main Results}
To evaluate the generalization capability of our method, we conduct experiments on standard zero-shot classification benchmarks like CLIP baseline and OnZeta. Table~\ref{tab:imagenet} presents the top-1 accuracy of our proposed LSA method on the ImageNet~\citep{imagenet_russakovsky2015imagenet} dataset using five CLIP backbones of varying complexity mentioned in the implementation section~\citep{clip_radford2021learning, resnet_he2016deep}. We categorize the results into three groups based on model scale and architecture: (1) convolution-based models (ResNet-50), (2) lightweight vision transformers (ViT-B/32 and ViT-B/16), and (3) large-scale transformers (ViT-L/14 and ViT-L/14@336px). To reduce variance caused by the stochastic arrival order of streaming data, each experiment across all datasets is repeated 5 times with different random permutations of the input stream. The final accuracy is reported as the average over the five runs.

For ResNet-50, LSA achieves a notable gain over the CLIP baseline, showing that our posterior adjustment significantly improves robustness even for weaker visual encoders. Among the mid-sized transformer models, ViT-B/16 achieves the highest absolute accuracy of $71.30\%$ and a relative gain of $2.49\%$, outperforming the CLIP baseline by $2.49\%$ and OnZeta by $0.49\%$, demonstrating strong alignment between LSA and models with moderate capacity. This also suggests that ViT-B/16 offers a favorable trade-off between performance and efficiency under online zero-shot settings. For larger architectures like ViT-L/14 and ViT-L/14@336px, the improvement margin narrows. These models already benefit from stronger generalization due to extensive pretraining and larger capacity, leaving less room for adaptation. Nevertheless, LSA still yields consistent gains, indicating its effectiveness even when the base model already exhibits strong generalization. These results demonstrate the robustness and architecture-agnostic compatibility of LSA. Moreover, unlike methods that require fine-tuning, LSA’s lightweight posterior adjustment easily incorporates into various backbones, enhancing performance without increasing computational burden. 

To further evaluate the generalization capability of LSA, we present results on the remaining 13 datasets in Table~\ref{tab:dataset}, spanning diverse tasks such as fine-grained recognition, scene understanding, texture classification, and satellite imagery. Experiments are conducted using both ResNet-50 and ViT-B/16 backbones. On average, our method surpasses OnZeta by $2.91\%$ with ResNet-50 and $2.02\%$ with ViT-B/16, demonstrating consistent improvements across architectures and domains. 

The ablation studies are provided in the supplementary material, including the influence of each component and the effect of varying the hyperparameter $\lambda_0$.

\section{Conclusion}
\label{sec:conclusion}
In this work, we addressed the challenging task of online ZSL with CLIP. This setting combines the semantic difficulty of ZSL with the practical constraints of online learning, where data arrive sequentially in a single pass and are not stored. Although prior methods focused on multimodal alignment and prompt engineering, they largely neglected the distribution mismatch between training and test data inherent to foundation models. To bridge this gap, we propose LSA, a novel framework that models online prediction as a domain adaptation problem under label shift. By dynamically estimating the evolving label distribution at test time, our method adjusts predictions without modifying CLIP's pre-trained weights or requiring access to labeled data. We evaluate the effectiveness of our method on multiple benchmark datasets. LSA achieves overall performance improvements across various settings and outperforms both OnZeta and CLIP baselines. The method achieves a new state of the art in online zero-shot classification with CLIP.

% \appendix


% %% For citations use: 
% %%       \cite{<label>} ==> [1]

% %%
% Example citation, See \cite{lamport94}.


\bibliographystyle{cas-model2-names} 
\bibliography{main}




% %% For numbered reference style

%   \textit{\LaTeX: a document preparation system},
%   Addison Wesley, Massachusetts,

% \end{thebibliography}

\end{document}